# Image Recognition of Oil Leakage Area Based on Logical Semantic Discrimination

Weiying Lin[1], Che Liu[1], Xin Zhang[1], Zhen Wei[1], Sizhe Li[2], Xun Ma[2]
Jinan Power Supply Company, Jinan, Shandong, China, 250000
Nanjing University Of Finance & Economics, Nanjing, Jiangsu, China, 210000

**Abstracts**：Implementing precise detection of oil leaks in peak load equipment through image analysis can significantly enhance inspection quality and ensure the system's safety and reliability. However, challenges such as varying shapes of oil-stained regions, background noise, and fluctuating lighting conditions complicate the detection process. To address this, the integration of logical rule-based discrimination into image recognition has been proposed. This approach involves recognizing the spatial relationships among objects to semantically segment images of oil spills using a Mask RCNN network. The process begins with histogram equalization to enhance the original image, followed by the use of Mask RCNN to identify the preliminary positions and outlines of oil tanks, the ground, and areas of potential oil contamination. Subsequent to this identification, the spatial relationships between these objects are analyzed. Logical rules are then applied to ascertain whether the suspected areas are indeed oil spills. This method's effectiveness has been confirmed by testing on images captured from peak power equipment in the field. The results indicate that this approach can adeptly tackle the challenges in identifying oil-contaminated areas, showing a substantial improvement in accuracy compared to existing methods.

## 0  Introduction.

With the continuous growth of power load, the operation and maintenance of peaking units has become an increasingly important issue in order to ensure the safe, stable and economic operation of the system. As the most commonly used peaking power equipment, hydropower units have the advantages of convenient start and stop, and easy synchronization when grid-connected. Since many devices in the unit are equipped with oil storage devices, oil leakage of the devices will bring potential safety hazards to the system. Therefore, maintenance managers need to find the oil leakage area in time and deal with it to ensure the normal operation of the corresponding equipment. In recent years, video surveillance technology in the hydropower station peaking power equipment condition monitoring task is widely used, if the computer vision technology to find equipment oil leakage and other abnormal phenomena can greatly reduce human resources consumption, while improving the quality of inspection [1-3].

At present, the most promising core technology in the field of computer vision is deep learning [4-6], whose advantage lies in the automatic extraction of key features to realize image recognition. Although this technology has achieved significant breakthroughs in the field of face recognition, security monitoring, etc., the identification of oil leakage in peaking equipment is still a difficult problem. The basic idea of oil spill image detection is to detect the oil stained area on the ground around the device, but compared with object recognition in daily life scenes, there are the following difficulties in oil stained area detection: 1) large differences in the shape of the area; 2) the complex background of the monitoring screen and the possible existence of staff, irrelevant objects and other interference; 3) changes in ambient lighting, smooth ground reflections, etc. The above difficulties make it difficult to recognize the oil spill area. The above difficulties make the recognition rate of the oil spill area not high.

In order to solve the problem of lighting changes, some studies[7] proposed to enhance the image brightness and contrast, and use the enhanced image for recognition. On this basis, to overcome the problem of shape difference, the idea of using difficult samples to train the model cyclically was proposed. Literature [8], on the other hand, uses the idea of converting the original image to HSV space and using image segmentation in the new space to determine the oil spill region. Literature [9] and [10], on the other hand, enhanced the image features using high-resolution and multi-scale fusion strategies, respectively, and learned the enhanced features using SSD model and YOLOv3 model. Despite the progress made in the above studies, relying entirely on image processing techniques to enhance model performance has significant limitations. Problems such as background interference and ground reflection cannot be solved by means of image processing, while the above factors are often the key to reduce the recognition accuracy.

Aiming at the above problems and the current research situation, this paper proposes an oil spill image recognition model based on logical semantic discrimination. Firstly, the image is enhanced using histogram equalization to improve the brightness and contrast of the region to be recognized. Then the semantic segmentation model is utilized to initially obtain a series of information such as object type, location, contour and so on. On this basis, this paper proposes an inter-object positional relationship recognition model and a logical discrimination model, the former of which determines the inter-object positional relationship by inputting information such as object type, position, and contour, and the latter of which determines the accuracy of the recognition results according to the set logical rules. Compared with the traditional image recognition model, the model proposed in this paper has the function of logical discrimination, which is of great significance in the field of power equipment recognition, which has rich prior knowledge. Meanwhile, compared with the existing research, the model proposed in this paper has stronger interpretability, and the basis of discrimination is more transparent, which is conducive to the use and promotion of the method in industrial scenarios.

# 1 Peak regulation power equipment oil leakage image recognition

This section introduces the algorithm framework for oil leakage image recognition of peak power equipment, which contains two links: image enhancement and semantic segmentation. The purpose of image enhancement is to reduce the influence of lighting, object reflection and other interference factors, and the purpose of semantic segmentation is to obtain the location information of the oil contaminated area.

**1.1 Image Enhancement of Oil Leakage Area**

The key to determine whether the peak power equipment is leaking oil is to identify the possible oil contamination area on the ground. The difficulty of oil contamination area detection is that the area has different shapes, the background is complex and there may be interfering objects, the light changes and there are reflections and other influencing factors. For the third point, image enhancement can be utilized to reduce the complexity of the problem. The core purpose of image enhancement is to enhance the image contrast, and the commonly used method is histogram equalization [11]. The main idea of this method is to approximate the histogram distribution of image pixel features into a uniform distribution, thus enhancing the contrast. For RGB images, the pixel features are adopted as Y-channel values after transformation into YCrCb space.

The flowchart of the method is shown in Fig. 1, in which the RBD (relative brightness difference) metric represents the relative brightness difference between the input image and the histogram equalized image, the RCD (relative contrast difference) metric represents the relative contrast difference between the input image and the histogram equalized image, the ASD (ASD) metric represents the relative contrast difference between the input image and the histogram equalized image, and the ASD (ASD) metric represents the relative contrast difference between the input image and the histogram equalized image. contrast difference, RCD (relative contrast difference) metric represents the relative contrast difference between the input image and the histogram equalized image, and ASD (average structural difference) metric represents the average structural difference between the input image and the histogram equalized image. After calculating the above indexes respectively, the BPS (brightness preservation score) function, OCS (optimum contrast score) function and ASD (detail preservation score) function are used to score respectively, and the scores of the three are used as the aggregated objectives for regular optimization (aggregated multiple objectives). optimization (aggregated multiple objectives regularization optimization) to get the final enhanced image.

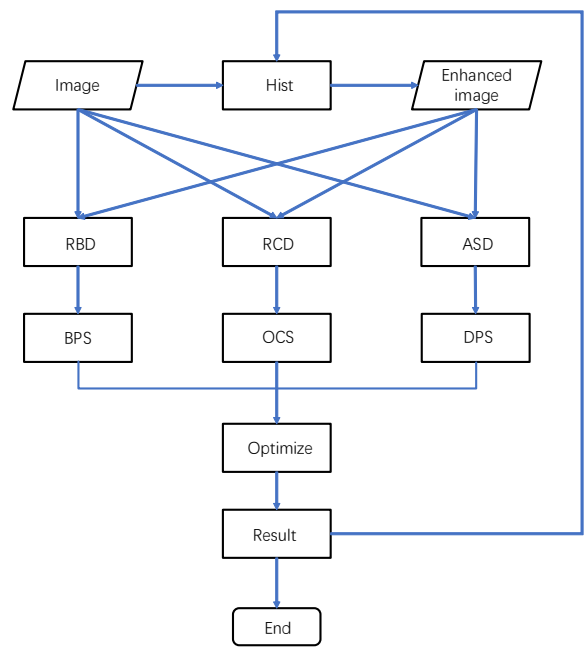

**Fig.1 Framework of Multipurpose Beta Optimized BiHE (MBOBHE) Model**

**1.2** Semantic Segmentation Model

In the field of computer vision, the semantic segmentation task is to further refine the recognition results on the basis of target detection, i.e., to achieve pixel-level front and rear view separation. For oiled area detection, semantic segmentation can obtain the pixel position information of the oiled area, which contains the important information of contour shape. The contour shape can well assist the model to determine whether the recognition result is an oil-contaminated area, and due to the slow process of oil spill, the shape and size of the oil-contaminated area exists in a development process, this feature can effectively help to identify the oil spill process in the live surveillance video.

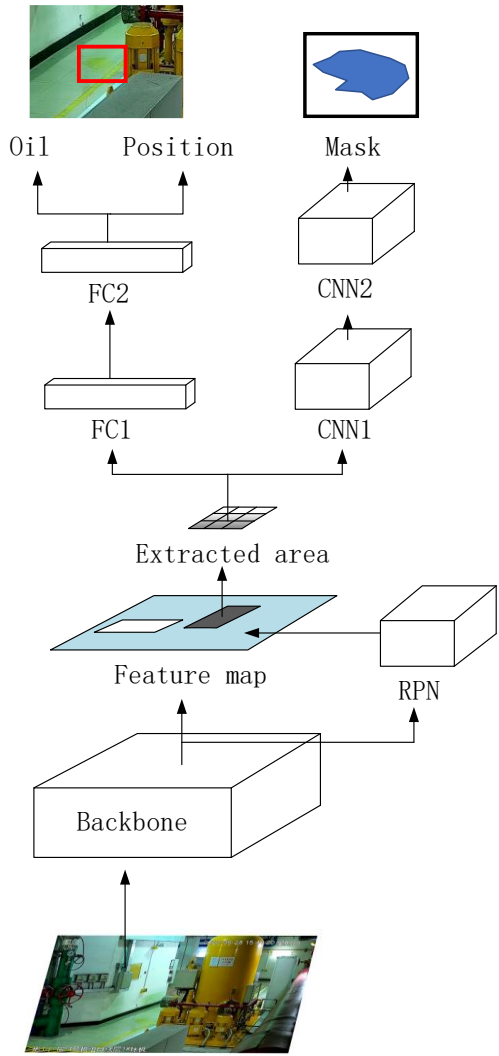

**Fig. 2 Network Architecture of Mask RCNN**

Mask RCNN network [12], as the most commonly used model in semantic segmentation, has high recognition accuracy and is easy to use and extend, so it is chosen as the basis of the recognition framework in this paper. The core idea of this method is to transform the semantic segmentation problem into an image region classification problem, and predict whether each pixel belongs to the foreground on the basis of classification. Figure 2 gives a schematic diagram of the structure of the network, which contains four parts: Backbone Network, Region Proposal Network, Classification Head and Mask Head. Among them, the feature extraction part of the original image is done by the base network, and the common model is ResNet network [13]. Here, in order to reduce data dependency, a pre-training model is usually used, i.e., a large dataset is used to train the network in advance, after which the network weights of the first layers are fixed, and the target dataset is used to fine-tune the latter layers. The candidate region where the target object is located is generated by the region candidate network (RPN) based on the extracted image features. After the candidate region is generated, the pooling layer (ROIAlign) extracts the corresponding features based on the location of the candidate region, which is used as input for the subsequent modules. The classification of the target object and the position correction are done by the region classification module, and the pixel location of the object, i.e., the mask, is done by the mask generation module. For the same object, the above two modules operate independently without interfering with each other. For different objects, the recognition process is carried out in parallel to ensure the efficiency of the algorithm.

## 1.3 Experimental Results

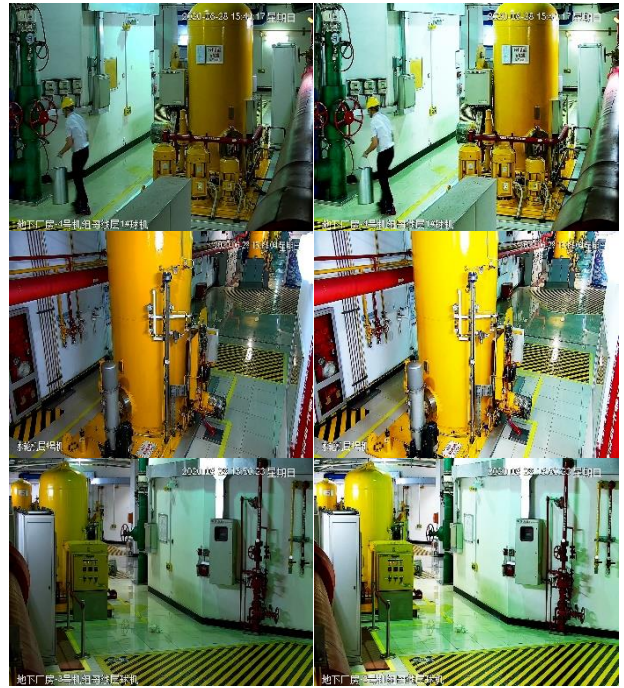

**(a) Original image     (b) Enhanced result**

**Fig. 3 Comparison between original and enhanced images**

The dataset used in this paper contains 237 images of FM power equipment. Labeling is done using the labelme annotation tool, i.e., the external contour of the target object is labeled using polygons, the data labeling format is polygon vertices, and the data storage format is a JSON file. After the dataset was created, 180 images were randomly selected as the training set and the remaining 57 images were used for testing. Cross-validation was used for training. The images contain two types: oil spill images and normal images, and the target objects contain oil spill areas, ground, and oil storage devices.

Figure 3 shows the comparison of image enhancement effects, where the left side is the original image and the right side is the enhanced image. It can be seen that the image enhancement algorithm used in this paper can well solve the interference problems caused by lighting, reflection, etc., and lay the foundation for the subsequent recognition algorithm. In order to alleviate the overfitting phenomenon, this paper uses data enhancement techniques.

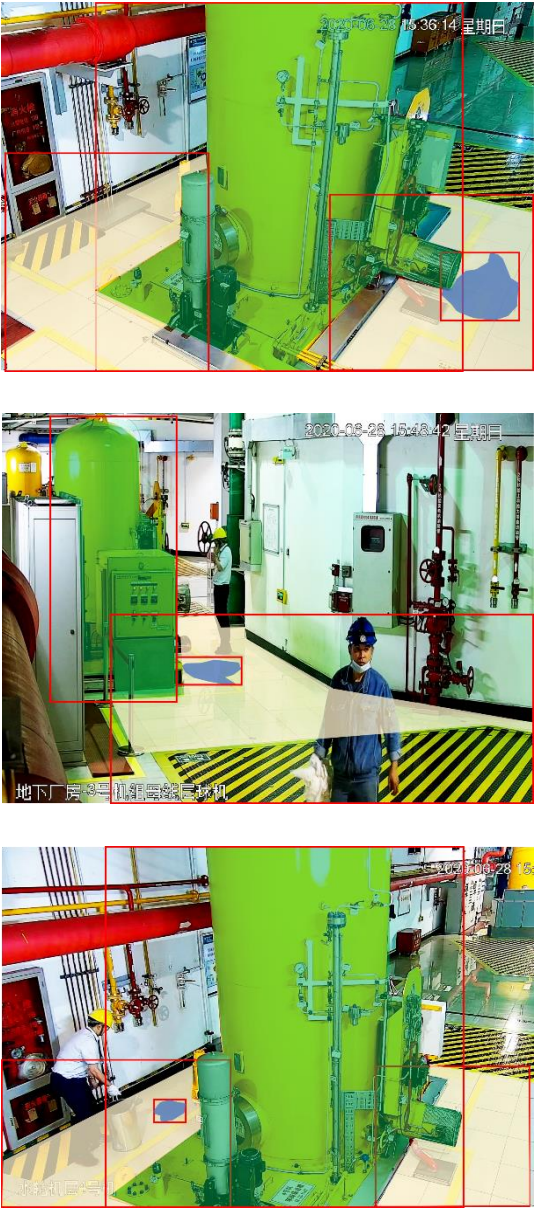

**Fig. 4 Results of image recognition**

Pre-trained models are also used, i.e., pre-trained models on large data and then fine-tuned on small datasets. Here the Mask RCNN model pre-trained on MS COCO dataset [14] is used and the base network is ResNet50 using NMS strategy. The network weight decay parameter is set to 0.0001, the momentum parameter is set to 0.9, and the learning rate is varied from 0.02 to 0.002 with the increase of iterations. In addition, a dropout layer [15] is added with a parameter of 0.2.The results of the image recognition are given in Fig. 4, where the object categories contain the oil spill area, the ground, and the storage device, and the information of both the bounding box and the mask is shown in the figure. It can be seen that different kinds of objects can be recognized effectively.

## 2 Logical semantic judgment

This section introduces the logical semantic judgment model framework in the oil leakage image of the peaking power equipment, which contains two links: the recognition of the positional relationship between objects and the logical discrimination. The purpose of inter-object positional relationship recognition is to capture the relevant information in the diagram to assist judgment, and the purpose of logical discrimination is to synthesize the results of image recognition and relevant auxiliary information based on logic.

**2.1 Inter-object positional relationship recognition**

Mask RCNN network does not solve the first two difficulties in the recognition of oiled areas, which have different shapes, complex backgrounds and possible interference objects. And the above problems precisely affect the accuracy of recognition to a large extent, so this paper carries out algorithmic improvement for the above difficulties. Inter-object position relation recognition is one of the difficulties in the field of computer vision [16,17], and the challenges it faces include the difficulty of data labeling, the loss of 3D to 2D position information, and the interference caused by object types. In this paper, since the object types and 3D position relations are relatively fixed, only the data labeling problem needs to be solved. Here the sample generation technique is used, i.e., the training samples are randomly generated using existing images of oiled areas, ground, oil storage devices, etc. For the generated samples, the positional relationship between objects is known and controllable, and it becomes possible to train the corresponding recognition module. The structure of the module is shown in Fig. 3, in which the input contains two sets of branches, dealing with the kind of location information and contour information, respectively, and the output is the result of inter-object relationship prediction, and the specific structure of the model is as follows:

$$\begin{cases} M_{ctr1} = conv_1(M_{mask}) \\ M_{ctr1} = pool(M_{ctr1}) \\ M_{ctr2} = conv_2(M_{ctr1}) \\ M_{ctr2} = pool(M_{ctr2}) \\ v_1 = f_{FC1}(v_{poi}, v_{cls}) \\ v_2 = f_{FC2}(v_1, M_{ctr2}) \\ y = \text{softmax}(v_2) \end{cases} \quad (1)$$

Where: $M_{mask}, v_{poi}, v_{cls}$ is the contour mask matrix, position vector and category vector, $conv_1, conv_2$ is the first convolution operation and the second convolution operation, $pool$ is the pooling operation, $f_{FC1}, f_{FC2}$ is the fully connected layer of the first and second layers, and $y$ is the prediction result.

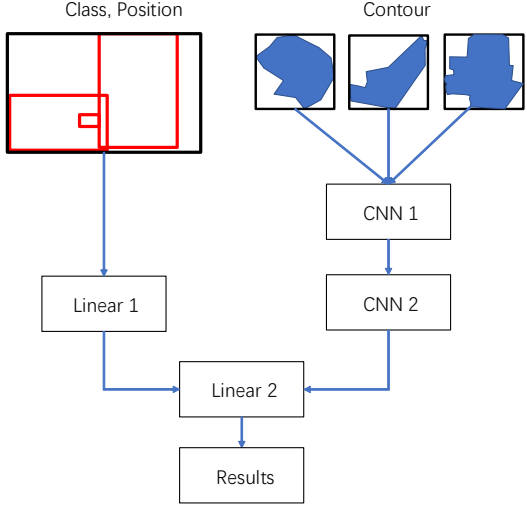

**Fig. 5 Object Positional Relation Detection Module**

**2.2 Logical Discriminative Module**

Compared with other image recognition models, the framework proposed in this paper has the ability of logic processing, so it can introduce a priori knowledge to effectively improve the performance of the model [18]. Here, the logic formula is introduced in detail with equation (2) as an example. The left side of equation (2) is the conclusion, and the right side is the premise, i.e., if all the premises are valid, the conclusion is valid. The single item in the formula is called predicate, which is used to describe the properties of the object or the relationship between the objects, such as "oil-contaminated area (A)" means that object A is an oil-contaminated area, and ABC refers to the object in the formula. Combining the above knowledge, it can be seen that the formula (2) indicates that "if the suspected oil-contaminated area A appears above the ground B, and there is an oil storage device C in its vicinity, it is concluded that A is an oil-contaminated area".

$$\text{Oil Area(A)} \leftarrow \text{Suspected Area(A)} \wedge \text{Ground(B)} \wedge \\ \text{Oil Storage Device(C)} \wedge \text{On(A,B)} \wedge \quad (2) \\ \text{around(A,C)}$$

In order to judge whether the premise is valid or not, it is necessary to judge whether each predicate in the premise is valid or not, and Eq. contains two kinds of predicates in total: 1) object attributes, such as "suspected oil-contaminated area, the ground, the oil storage device", and the validity of this part of the predicate is given by the semantic segmentation module; 2) inter-object relations, such as "above, nearby", and the relationship between objects, such as "above, near". above, in the vicinity", this part of predicate establishment is given by the inter-object location relationship recognition module.

Since the discriminative results given by the above modules are probabilistic values, a fuzzy logic inference module is designed here, which contains three sub-modules: logical and $\wedge$, logical or $\vee$ and logical not $\neg$, of which the three calculations are as follows:

$$\begin{cases} y \leftarrow x_1 \wedge x_2 & \Rightarrow y = b_1 x_1 + b_2 x_2 + c \\ y \leftarrow \neg x & \Rightarrow y = 1 - x \\ y \leftarrow x_1 \vee x_2 & \Rightarrow y = \max(x_1, x_2) \end{cases} \quad (3)$$

Where: $x_1, x_2, x, y \in [0,1]$ cis the fuzzy logic variable and $b_1, b_2, c$ is the parameter to be solved. Through the above fuzzy logic inference module, the premise can be judged to be established situation.

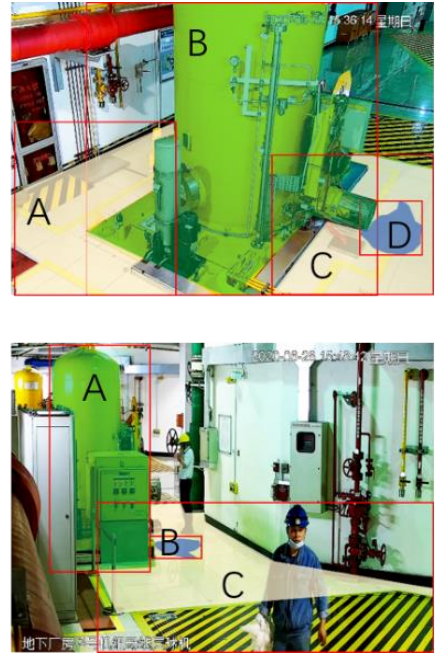

**Fig. 6 Object Positional Relation Detection Results**

**2.3 Experimental results**

The kind, position, and outline information in the image recognition result is used as the input of the module of positional relationship recognition between objects, and the output of the module is the result of relationship prediction. The objects are paired two by two at the time of input. The size of the original contour information is $28 \times 28 \times 1$, after the first layer convolution becomes $14 \times 14 \times 256$, after the second layer convolution becomes $3 \times 3 \times 256$. The number of neurons in the first fully-connected layer is 1024, and the number of neurons in the second fully-connected layer is 256. Fig. 6 illustrates some of the results, where ABCD denotes different objects, and the final prediction results contain three categories: "on top",

"nearby ", and "Other". Among them, "above" is a directional relationship, and the order of objects cannot be replaced, while "near" is a non-directional relationship. It can be seen that the module proposed in this paper can recognize inter-object relationships well.

On this basis, logical semantic judgment is performed, and the rules used in the process are shown below. Gradient descent method is used for parameter fetching in the fuzzy logic inference module, and cross-entropy is used for the error function, and the final prediction result is the probability of oil leakage from the peaking power equipment in this figure. The parameter values corresponding to different rules are shown in Table 1.

$$\text{Oil Area(A)} \leftarrow \text{Suspected Area(A)} \wedge \\ \text{Ground(B)} \wedge \text{on(A,B)} \quad (4)$$

$$\text{Oil Area(A)} \leftarrow \text{Suspected Area(A)} \wedge \\ \text{Oil Storage Device(B)} \wedge \text{around(A,B)} \quad (5)$$

$$\text{Oil Area(A)} \leftarrow \text{Suspected Area(A)} \wedge \text{Ground(B)} \wedge \\ \text{Oil Storage Device(C)} \wedge \text{On(A,B)} \wedge \quad (6) \\ \text{around(A,C)}$$

**Tab.1 Parameter values in different rules**

| Rule | Weight vector $[b_1, b_2, \cdots, c]$ |
|---|---|
| (4). | [0.645, 0.181, 0.162, 0.012] |
| (5) | [0.390, 0.323, 0.247, 0.040] |
| (6) | [0.417, 0.080, 0.297, 0.062, 0.124, 0.020] |

## 3 Model Analysis

In this section, the model is analyzed in detail, including the validation of the reasonableness of the input information, structural design, parameter values, and other stages, while comparing the methods in the related literature to prove the superiority of the model proposed in this paper.

### 3.1 The model proposed in this paper

The model proposed in this paper contains four modules, namely, image enhancement, semantic segmentation, recognition of positional relationship between objects and logical semantic judgment, and each module will be analyzed separately here. For the image enhancement module, its effect on semantic segmentation is tested here. There are two main aspects to consider in judging the accuracy of semantic segmentation: object type and location, where the location is judged by the IoU (Intersection over Union) metric, i.e., the degree of overlap between the real rectangular bounding box and the predicted result. In order to fully reflect the model performance, the recognition accuracy AP (Average Precision) under the thresholds of different overlapping degree is given here, AP50 corresponds to a threshold of 0.5, AP75 corresponds to a threshold of 0.75, and mAP denotes the average value under the threshold change from 0.5 to 0.95. The upper half of the figure shows the segmentation results corresponding to the original image, and the lower half shows the results corresponding to the enhanced image. It can be seen that the recognition accuracy of the oil spill area increases significantly, while the recognition accuracy of the ground and the oil storage device does not change much. The main purpose of this module is to enhance the brightness and contrast of the image, and the enhancement is more obvious for the recognition of the oil spill region.

Next is the inter-object location relationship recognition module, where the effect of different input information on the accuracy rate is first tested, as shown in Table 2. It should be noted that the evaluation metric used in this paper is the F1 score, which is defined as the reconciled average of accuracy and recall. Compared to the accuracy rate, the metric also considers the recall rate to illustrate the model's ability to detect real samples.

**Tab.2 Object relation classification accuracy under different types of input**

| Model input | Above | Nearby | Other | Total |
|---|---|---|---|---|
| Location Information | 0.480 | 0.550 | 0.621 | 0.550 |
| Location information + type information | 0.524 | 0.571 | 0.571 | 0.583 |
| Location information + Type information + Profile information | **0.601** | **0.651** | **0.801** | **0.684** |

The model performance under different combinations of input information is validated separately in Table 2. It can be seen that the category and silhouette information can be used as an aid to the location information to enhance the model recognition accuracy. This is because the corresponding positional relationships between different categories of objects are distributed differently, while the contour information as a supplement to the positional information can also play a role in helping the model judgment. In addition, for the location relationships between different objects, "above" has the lowest recognition accuracy, followed by "near", and the highest is "other". This is due to the fact that different relationships have different recognition difficulties, which depend on the accuracy of the input information as well as the differences between the relationships. "Above" often corresponds to the oil spill area and the ground, which are difficult to accurately determine the location information of these two types of objects, and they are easily confused with "near", so it is more difficult to recognize them. However, as can be seen from Table 2, by adding the type information and contour information, the recognition accuracy of this relationship is significantly improved, which proves the effectiveness of the method proposed in this paper.

For the object positional relation classification module designed in this paper, a hyperparameter test is conducted to determine the best parameter selection. The test parameters are the number of convolutional layers and the number of linear layers, the test object is the training set and the test set, the test evaluation index is the recognition accuracy rate. It can be seen that with the increase of the number of convolutional layer layers, the recognition accuracy rate in the training set increases gradually, but the accuracy in the test set increases first and then decreases. This is due to the high dimensionality of the contour information, and when the number of convolutional layers is small, there is a lack of information integration, and the accuracy is not high in the case of a small number of samples. As the number of layers increases, the degree of information integration is higher and higher, but at the same time, the number of model parameters is also increasing, and the overfitting phenomenon is gradually serious. Therefore, it is necessary to reasonably select the number of convolutional layers, and the optimal parameter obtained here is 2. Similarly, we can analyze the influence of the number of linear layers on the recognition accuracy, and obtain the optimal value of 2 for the number of linear layers.

After the testing of a single module, the effect of different modules on the final recognition accuracy is verified here, as shown in Table 3. The table involves different model structures, in which "image recognition" means using AlexNet model [19] to classify the whole picture directly, and the other three model structures, i.e., by using the different modules proposed in this paper, to ultimately determine whether there is an oil leakage area in the picture. It can be seen that the complete framework proposed in this paper improves the recognition accuracy by 0.315 compared to the most basic image recognition model, which is significant for industrial level applications. At the same time, image enhancement, positional relationship recognition and logical semantic judgment have significantly improved the semantic segmentation results, which is because the above modules have well solved the difficulties in the task of oil spill image recognition (different shapes of the region, complex backgrounds and possible interference objects, lighting changes and reflection and other influencing factors), which fully explains the significance of designing the corresponding model improvement strategies for specific application scenarios. The results fully demonstrate the significance of designing appropriate model improvement strategies for specific application scenarios.

**Tab.3 Accuracy of different model structures**

| Model | Normal picture | Oil leakage picture | Total |
|---|---|---|---|
| Image Recognition | 0.521 | 0.415 | 0.468 |
| Semantic Segmentation | 0.654 | 0.510 | 0.582 |
| Image Enhancement + Image Recognition | 0.682 | 0.584 | 0.633 |
| Image Enhancement + Image Recognition + Positional Relationship Recognition + Logical Semantic Judgment | **0.810** | **0.755** | **0.783** |

### 3.2 Comparison of Related Work

In order to reflect the superiority of the method proposed in this paper, it is compared here with related works, including the difference method model [8], the loop training method model [7] and the improved SSD algorithm [9]. The core idea of the difference method model is to compare the HS color histograms of the target region, and the algorithm process is as follows: obtaining the difference value of the grayscale image, segmenting the image by using the OTSU method [20], and transforming the segmented anomalous region into the HSV space for judgment. The core idea of the cyclic training method model is to train the model cyclically for difficult samples [21], and the algorithm flow is as follows: augment the original image, and keep mining difficult samples to join the training set until the training is completed. The core idea of the improved SSD algorithm is to add multi-resolution features to improve the model performance, and the algorithm process is: randomly mask part of the region, add the generated masked image to the training set, and at the same time, fuse the features under multi-resolution, and use the SSD network to determine whether there is an oil leakage region.

Table 4 gives the recognition accuracy corresponding to different models, it is easy to see that the proposed model recognition accuracy is significantly better than the remaining three models. For the difference method model, since the model is based on the strong assumption that the difference between the before and after images is only the oil leakage area, the model has low robustness, poor migration, and also puts forward high requirements for the resolution of the input image, which is unsuitable for the scenario of monitoring oil leakage of peaking power supply equipment under surveillance video. For the cyclic training method model, although the model improves the accuracy to a certain extent by introducing difficult sample mining, but due to the lack of correction means such as logical judgment, the model performance is easy to be interfered by the scene personnel, similar background objects, and the stability is not strong. Although the improved SSD model introduces data enhancement technology (random masking) and multi-resolution feature fusion, it still does not target the difficulties in oil spill image discrimination, such as different shapes of the region, complex backgrounds and the possible existence of interfering objects, changes in lighting and the existence of reflections and other influencing factors, so the model performance is limited to improve the identification accuracy of the model proposed in this paper, there is a large gap between the identification accuracy of the model. In summary, as this paper addresses the difficulties of peak power equipment oil leakage image recognition in this specific scenario, it proposes a series of effective solutions such as image enhancement, positional relationship recognition between objects, logical semantic discrimination, etc., and its recognition accuracy is significantly higher than that of similar methods. Therefore, this method is of reference significance for the task of equipment state discrimination under the conditions of image shooting and video monitoring in the electric power industry.

**Tab.4 Accuracy of different models**

| Model | Normal image | Oil leakage picture | Total |
|---|---|---|---|
| Differential model | 0.621 | 0.351 | 0.486 |
| Cyclic training method model | 0.731 | 0.710 | 0.721 |
| Improved SSD algorithm | 0.691 | 0.674 | 0.683 |
| Logical semantic judgment | **0.810** | **0.755** | **0.783** |

## 4  Conclusion

In this paper, we propose an oil leakage image recognition method for peaking equipment based on logical semantic judgment, which utilizes logical relations to improve the recognition accuracy in limited training samples, which is conducive to realizing the automation of equipment condition monitoring and improving the quality of inspection. The main conclusions are as follows:

(1) The original image is enhanced based on histogram equalization technology, which is conducive to subsequent image recognition; the location and contour information of the oil storage device, the ground, and the suspected oil contaminated area are initially obtained using Mask RCNN network.

(2) Aiming at the problems of background interference and ground reflection in the identification of oil contaminated areas, a module for identifying the positional relationship between objects is proposed, which determines the positional relationship of different objects according to their positional and contour information, and the identification accuracy is close to 70%.

(3) A logical discrimination module is proposed to determine whether a suspected area belongs to an oil spill area based on the positional relationship between objects combined with logical expressions, which increases the model recognition accuracy to nearly 80%, far exceeding the existing methods, and the overall framework has the ability to incorporate the a priori knowledge, and the model performance has the potential to be further improved.

## References:

[1] Mao, Xianyin, Ma Xiaohong, and Feng Junkuan. "Intelligent defect identification system of power equipment based on UAV patrol image technology." China Energy and Environmental Protection, vol. 43, no. 7, 2021, pp. 225-230.

[2] Cui, Zhuo, He Quan-wei, Yu Yi-sheng, et al. "Power equipment image recognition based on video surveillance." Information Technology, no. 6, 2021, pp. 27-32, 39.

[3] Chen, Peng, and Qin Lunming. "Infrared Image Recognition of Power Equipment Based on Deep Learning." Journal of Shanghai University of Electric Power, vol. 37, no. 3, 2021, pp. 217-220, 230.

[4] Redmon, J., Divvala, S., Girshick, R., et al. "You only look once: Unified, real-time object detection." Proceedings of the IEEE Conference on Computer Vision and Pattern Recognition, 2016, pp. 779-788.

[5] Girshick, R. "Fast r-cnn." Proceedings of the IEEE International Conference on Computer Vision, 2015, pp. 1440-1448.

[6] Liu, W., Anguelov, D., Erhan, D., et al. "SSD: Single shot multibox detector." European Conference on Computer Vision, Springer, Cham, 2016, pp. 21-37.

[7] Bao, Weichao, Gu Li, He Jinsong, et al. "Transformer Oil Leakage Detection Based on Loop Training Method." Journal of Computer-Aided Design & Computer Graphics, vol. 33, no. 03, 2021, pp. 431-438.

[8] Baoguo, Dong. "Detection of Transformer Oil Leakage Based on Image Processing." Electric Power Construction, vol. 34, no. 11, 2013, pp. 121-124.

[9] Feng, Tingyou, Cai Chengwei, Tian Ji, et al. "An Improved SSD Algorithm for Oil Leakage Detection of Power Plant Equipment." Computer Measurement & Control, 2021, pp. 1-8.

[10] GONG Yu, LU Chuande, FU Yanqing, et al. "Detection of Water and Oil Leakage in Production Area of Power Plant Based on Improved YOLOv3 Model." Guangdong Electric Power, vol. 34, no. 06, 2021, pp. 55-64.

[11] Hum, Y. C., Lai, K. W., & Mohamad Salim, M. I. "Multiobjectives Bihistogram Equalization for Image Contrast Enhancement." Complexity, vol. 20, no. 2, 2014, pp. 22-36.

[12] He, K., Gkioxari, G., Dollár, P., et al. "Mask r-cnn." Proceedings of the IEEE International Conference on Computer Vision, 2017, pp. 2961-2969.

[13] He, K., Zhang, X., Ren, S., et al. "Deep residual learning for image recognition." Proceedings of the IEEE Conference on Computer Vision and Pattern Recognition, 2016, pp. 770-778.

[14] Lin, T. Y., Maire, M., Belongie, S., et al. "Microsoft coco: Common objects in context." European Conference on Computer Vision, Springer, Cham, 2014, pp. 740-755.

[15] Srivastava, N., Hinton, G., Krizhevsky, A., et al. "Dropout: a simple way to prevent neural networks from overfitting." The Journal of Machine Learning Research, vol. 15, no. 1, 2014, pp. 1929-1958.

[16] Xiong, Siheng, et al. "Object recognition for power equipment via human‐level concept learning." IET Generation, Transmission & Distribution 15.10 (2021): 1578-1587.

[17] Liu, M., Wang, H., Li, Y., et al. "Research on Visual Relation Detection Based on Computer Vision." 2020 3rd International Conference on Advanced Electronic Materials, Computers and Software Engineering (AEMCSE), IEEE, 2020, pp. 342-345.

[18] Yang, Yuan, et al. "Temporal Inductive Logic Reasoning." arXiv preprint arXiv:2206.05051 (2022).

[19] Krizhevsky, A., Sutskever, I., Hinton, G. E. "Imagenet classification with deep convolutional neural networks." Advances in Neural Information Processing Systems, vol. 25, 2012.

[20] Bangare, S. L., Dubal, A., Bangare, P. S., et al. "Reviewing Otsu's method for image thresholding." International Journal of Applied Engineering Research, vol. 10, no. 9, 2015, pp. 21777-21783.

[21] Shrivastava, A., Gupta, A., Girshick, R. "Training region-based object detectors with online hard example mining." Proceedings of the IEEE Conference on Computer Vision and Pattern Recognition, IEEE Computer Society Press, 2016, pp. 761-769.

[22] Dai, B., Zhang, Y., Lin, D. "Detecting visual relationships with deep relational networks." Proceedings of the IEEE Conference on Computer Vision and Pattern Recognition, 2017, pp. 3076-3086.

[23] Dong, Baoguo. "Detection of transformer oil leakage based on image processing." Electric Power Construction, vol. 34, no. 11, 2013, pp. 121-124.